\documentclass[letterpaper, 10 pt, conference]{ieeeconf}  

\IEEEoverridecommandlockouts                              

\title{\LARGE \bf
Aegis-VLA: Reliability-Gated Action Refinement \\
via a Trajectory World Model for Dynamic Object Manipulation
}

\author{Lennox Wei$^{1}$, Jiaqi Yao$^{1}$, Shixin Sun$^{1}$, Yuanzheng Zhang$^{2}$, \\ Jianhua Han$^{3}$, Meng Tian$^{3,4}$, Shaohui Liu$^{1*}$, Wangmeng Zuo$^{1}$
\thanks{*Corresponding author: Shaohui Liu (e-mail: shliu@hit.edu.cn). }
\thanks{$^{1}$Lennox Wei, Jiaqi Yao, Shixin Sun, Shaohui Liu and Wangmeng Zuo are with School of Computer Science and Technology, Faculty of Computing, Harbin Institute of Technology, Harbin 150001, China.}%
\thanks{$^{2}$Yuanzheng Zhang is with University of Southern California, Los Angeles, CA 90089, USA.}
\thanks{$^{3}$Jianhua Han and Meng Tian are with Yinwang Intelligent Technology Co., Ltd., Shanghai 200135, China.}
\thanks{$^{4}$Meng Tian is with National University of Singapore, Singapore 119077, Singapore.}
}

\usepackage{graphicx} 
\usepackage{amsfonts, amssymb, amsmath, booktabs, multirow, colortbl}
\usepackage{xcolor}
\usepackage{booktabs}
\usepackage{caption}
\usepackage{array}
\usepackage{cite}
\usepackage[justification=raggedright]{caption}

\usepackage[ruled,linesnumbered]{algorithm2e}
\usepackage{amsmath,amssymb,amsfonts}
\usepackage{pifont} 
\usepackage{array} 
\usepackage{pifont}
\usepackage{tabularx}
\usepackage{booktabs} 
\newcolumntype{C}{>{\centering\arraybackslash}X} 
\usepackage{makecell} 
\usepackage{pifont}   
\usepackage{tabularx}
\usepackage{makecell}
\usepackage{pifont}   
\usepackage{bm}       
\usepackage{booktabs}
\usepackage{multirow}
\usepackage{graphicx}
\usepackage{amsmath}
\usepackage[format=plain,justification=justified,singlelinecheck=false]{caption}
\usepackage{amsmath,amssymb}
\usepackage{cite}
\usepackage{url}
\usepackage{booktabs,array}
\usepackage{cite}                               
\usepackage{amsmath,amssymb,amsfonts}           
\usepackage{graphicx}                           
\usepackage[table]{xcolor}                      
\usepackage{booktabs}                           
\usepackage{multirow}                           
\usepackage{makecell}                           
\usepackage{tabularx}                           
\usepackage{adjustbox}                          
\usepackage{pifont}                             
\usepackage{bbm} 
\usepackage{graphicx,booktabs,tabularx,amsmath,amssymb}
\usepackage{xcolor,colortbl}
\usepackage{hyperref}

\definecolor{mygreen1}{HTML}{32CD32} 
\definecolor{mygreen2}{HTML}{90EE90} 
\definecolor{mygreen3}{HTML}{E2F0D9} 
\definecolor{myorange1}{HTML}{F4B183} 
\definecolor{myorange2}{HTML}{FBD4B4} 
\definecolor{myorange3}{HTML}{FFFACD} 
\definecolor{mypurple}{HTML}{E6E6FA}

\begin{document}

\maketitle
\thispagestyle{empty}
\pagestyle{empty}

\begin{abstract}
Dynamic object manipulation requires maintaining spatial and temporal alignment with targets that continue to move during inference and execution. Although future prediction can inform action correction, an accurate forecast does not establish whether modifying an existing action would be beneficial. We present Aegis-VLA, a world-model-guided vision-language-action (VLA) framework that generates predictive action corrections and separately determines whether and how strongly to apply them. A Trajectory World Model combines kinematic extrapolation with learned residual dynamics to predict object motion and estimate positional uncertainty. Discrepancy-Guided Action Refinement uses differences between history-only and base-action-conditioned end-effector forecasts to propose bounded arm corrections. A Reliability Gate learns when to admit corrections based on demonstration-relative improvements over base actions, and regulates their application during inference using predictive uncertainty and anticipated object--end-effector proximity. Rejected corrections preserve the corresponding base actions. Aegis-VLA introduces only approximately 2.45M trainable parameters. On the simulated Dynamic Object Manipulation (DOM) benchmark, Aegis-VLA achieves 56.72\% success over 1,800 evaluation trials, exceeding our reproduced DynamicVLA baseline of 45.44\% by 11.28 percentage points.
\end{abstract}

\section{Introduction}
\label{sec:introduction}

Manipulating a moving object requires the robot and target to meet at the right place and time~\cite{dbctfp,gem,dynamicvla}. A grasp that is geometrically appropriate at planning time can become ineffective before contact if the object continues moving. The same observed position can demand different actions depending on whether the target is approaching, receding, or changing direction. Partial visibility obscures motion, while perception, inference, and physical execution introduce delays during which the scene evolves. Dynamic manipulation therefore requires maintaining spatial and temporal alignment between robot actions and target motion.

Pretrained vision-language-action (VLA) and generalist robot policies provide increasingly capable foundations for language-conditioned manipulation~\cite{rt2,openvla,pizero}. Recent methods further incorporate motion history~\cite{tracevla}, improve inference efficiency~\cite{smolvla}, and coordinate action generation with execution~\cite{chunk,vlash}. 
Temporal observations, continuous inference, and temporally aligned action streaming enable VLA policies to account for aspects of target motion and execution timing~\cite{dynamicvla}.
We therefore ask how predictive evidence can justify modifying a base action that may already be appropriate.

\begin{figure}
    \vspace{+0.15cm}
    \centering
    \includegraphics[width=0.99\linewidth]{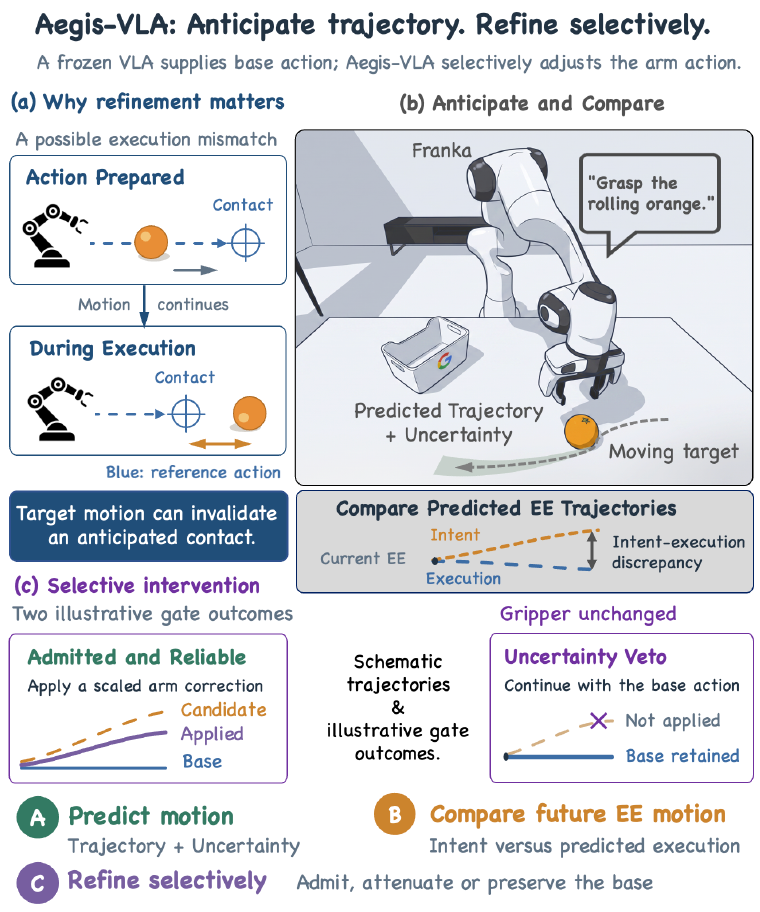}
    \vspace{-0.35cm}
    \caption{Motivation of Aegis-VLA. Target motion can invalidate an anticipated contact during execution. Aegis-VLA predicts target motion and uncertainty, compares end-effector forecasts, and selectively applies scaled arm corrections or retains the base action.}
    \label{left}
    \vspace{-0.55cm}
\end{figure}

World models offer information about how a scene may evolve beyond the latest observation. Video and representation prediction can support policy learning and planning~\cite{gr1,vpp,worldvla}, while recent approaches explicitly incorporate target motion or predicted future observations into dynamic manipulation~\cite{puma,dynamicwam,ahead}. World-model-based approaches also evaluate or revise action proposals~\cite{dreamtrajectory,dynawm}. However, predicting target motion alone does not establish whether applying a proposed correction would improve execution compared with leaving the base action unchanged.

As illustrated in Fig.~\ref{left}(a), target motion can invalidate an anticipated contact during execution. A forward adjustment may improve alignment when the base action trails the target. If the base action already compensates for that motion, however, an additional adjustment may worsen alignment even with an accurate forecast. An apparently useful correction may also depend on an unreliable forecast. Useful refinement must therefore assess whether a proposed correction improves on the base action and whether the motion forecast is sufficiently reliable to justify the correction.

\begin{figure*}
    \vspace{+0.15cm}
    \centering
    \includegraphics[width=0.99\linewidth]{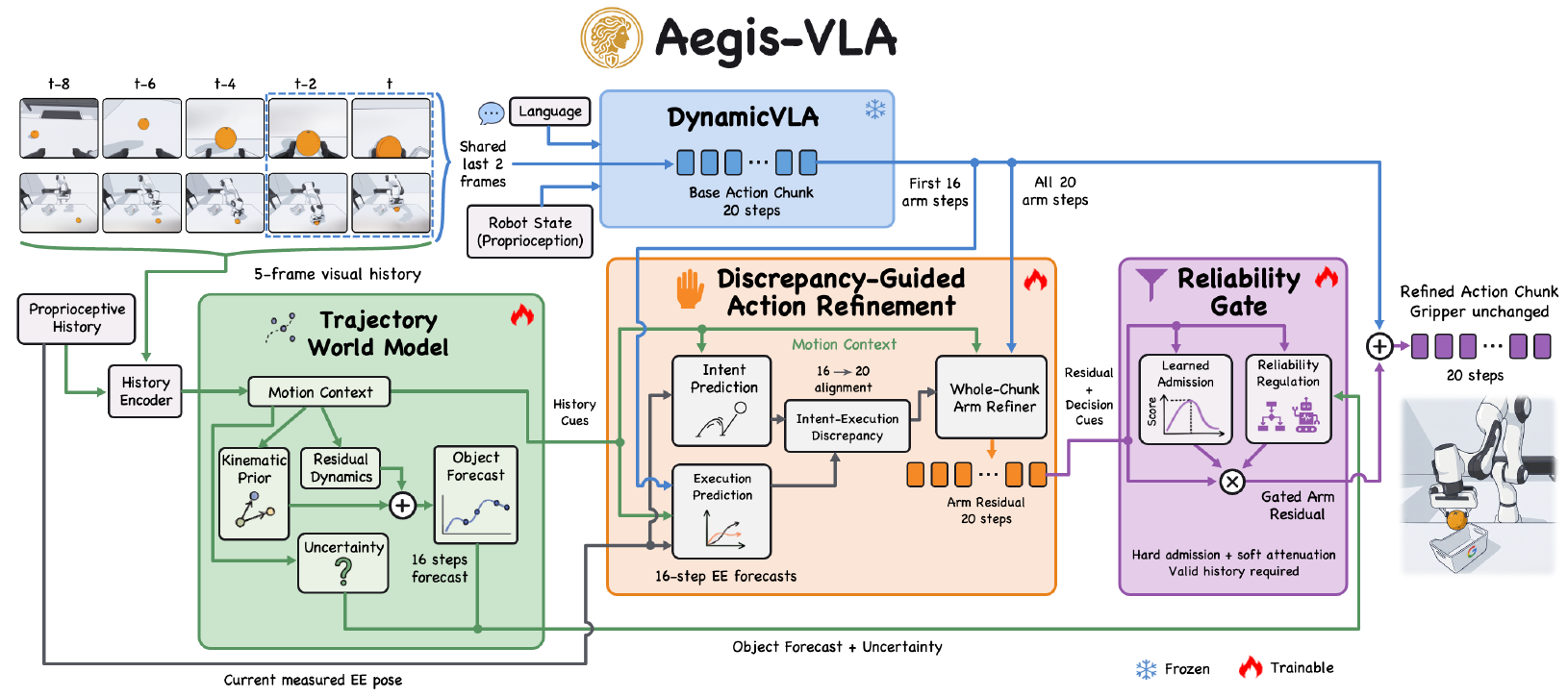}
    \caption{Framework of Aegis-VLA. A frozen VLA backbone generates base actions, while the Trajectory World Model supplies motion context, object forecasts, and positional uncertainty. Discrepancy-Guided Action Refinement produces bounded arm residuals, which the Reliability Gate regulates before execution.}
    \label{overview}
    \vspace{-0.55cm}
\end{figure*}

Residual learning provides a way to adapt an existing controller or policy~\cite{residualrl,policydecorator}, including observation-conditioned per-step action correction~\cite{a2c2} and uncertainty-guided corrective exploration~\cite{uncertaintyresidual}. However, freezing backbone weights does not by itself preserve task performance once residuals alter the executed actions. Aegis-VLA builds on this residual formulation by separating correction generation from admission through the three coordinated modules in Fig.~\ref{overview}.


Our contributions are summarized below:
\begin{itemize}
    \item \textbf{Trajectory World Model.} We combine kinematic extrapolation, learned residual dynamics, and positional uncertainty estimation to provide shared motion context for action refinement and predictive evidence for regulating corrections under uncertain target motion.
    \item \textbf{Discrepancy-Guided Action Refinement.} We generate bounded arm corrections from position and orientation discrepancies between history-only and base-action-conditioned end-effector forecasts, linking each proposal to the base action it modifies.
    \item \textbf{Reliability Gate.} We separate correction proposal from adoption by learning admission from demonstration-relative improvement of each candidate over its base action, while uncertainty and proximity regulate application at inference.
\end{itemize}

Aegis-VLA introduces only approximately 2.45M trainable parameters. On the simulated Dynamic Object Manipulation (DOM) benchmark~\cite{dynamicvla}, Aegis-VLA achieves 56.72\% success over 1,800 evaluation trials, exceeding our reproduced DynamicVLA baseline of 45.44\% by 11.28 percentage points. Ablations demonstrate that structured refinement already improves the backbone, while learned admission and reliability regulation further improve on an independently trained ungated alternative.

\section{Related Work}
\label{sec:related_work}

\subsection{Dynamic Manipulation and Responsive Execution}

Vision-language-action (VLA) models and generalist robot policies provide reusable foundations for manipulation~\cite{rt2,openvla,octo,pizero,pizerofive}, leveraging temporally coherent action sequences~\cite{act,diffusionpolicy} and efficient action generation~\cite{openvlaoft,smolvla}. Dynamic manipulation additionally requires motion-aware perception and control~\cite{dbctfp,gem,tracevla,contextvla} together with responsive execution~\cite{chunk,rtvla,vlash}.


DynamicVLA~\cite{dynamicvla} combines temporal observations, continuous inference, and temporally aligned action streaming. Whereas A2C2~\cite{a2c2} corrects each action from recent observations and base-action context, Aegis-VLA generates a bounded residual chunk from discrepancies between two predicted end-effector trajectories.

\subsection{World Models for Anticipatory Manipulation}

Video-based policies and world models connect predictive representations to action generation~\cite{unipi,gr1,vpp,genieenvisioner,worldvla,hifvla}. For dynamic manipulation, PUMA~\cite{puma} combines historical optical flow with auxiliary prediction of future object-centric features, while DynamicWAM~\cite{dynamicwam} conditions joint video and action generation on flow and kinematic descriptors.

AHEAD~\cite{ahead} supplies predicted visual tokens to a frozen VLA and uses uncertainty to adapt the rollout horizon. Aegis-VLA instead predicts object-coordinate trajectories and positional uncertainty from dual-view RGB and proprioceptive history to support action-space refinement. Forecasting supervision shapes the shared motion context, while explicit trajectories support proximity estimation. Its forecast horizon is fixed, and uncertainty regulates corrections.


\subsection{Prediction-Guided and Selective Action Refinement}

Residual policy learning adapts existing controllers through additive corrections~\cite{residualrl,resip,policydecorator}. Uncertainty-aware residual reinforcement learning regulates exploration using base-policy uncertainty~\cite{uncertaintyresidual}. Aegis-VLA instead uses uncertainty in predicted object motion to constrain correction application during inference.

DreamTrajectory~\cite{dreamtrajectory} ranks sampled action chunks, including the base proposal, whereas DynaWM~\cite{dynawm} regenerates complete chunks conditioned on a frozen VLA's base actions and observation history. Unlike chunk-level ranking or regeneration, Aegis-VLA explicitly learns admission for individual corrections within a single bounded residual chunk. The residuals are generated directly from discrepancies between history-only and base-action-conditioned end-effector forecasts, without candidate search or additional sampling of action chunks from the backbone.

\section{Methodology}
\label{sec:aegis_method}

\subsection{Overview}
\label{sec:aegis_overview}

Fig.~\ref{overview} presents the overall architecture of Aegis-VLA. 
At observation index $t$, Aegis-VLA's predictive refinement modules receive a history $\mathcal H_t$ containing five external-view and wrist-view RGB pairs and aligned end-effector poses $\mathbf s_j\in\mathbb R^6$. Here, $t$ and $j$ are discrete observation indices, $\mathcal H_t$ denotes the visual--proprioceptive history, $j\in\{t-8,t-6,t-4,t-2,t\}$ selects its entries, and $\mathbf s_j$ contains Cartesian position in meters and Euler orientation in radians.
The observations are sampled at frequency $f_{\mathrm{s}}=25\,\mathrm{Hz}$, 
and the five entries span $0.32\,\mathrm{s}$. 
Aegis-VLA uses a frozen DynamicVLA~\cite{dynamicvla} policy as its action-generation backbone, which receives the final two image pairs from the history, at offsets $-2$ and $0$, together with the language instruction and proprioceptive inputs.
A binary history-validity flag $\chi_t$ is set to one when all five required history entries are available.

The backbone produces a chunk of arm actions $\mathbf U_t^{\mathrm{base}}=[\mathbf u_{t,i}^{\mathrm{base}}]_{i=1}^{K}\in\mathbb R^{K\times6}$ and gripper actions $g_{t,i}^{\mathrm{base}}$. Here $K$ is the chunk length, $i$ indexes an action step, $\mathbf u_{t,i}^{\mathrm{base}}\in\mathbb R^6$ is an arm action, $g_{t,i}^{\mathrm{base}}\in\mathbb R$ is its scalar gripper action, the superscript $\mathrm{base}$ denotes the backbone output, and bracketed indexed vectors are stacked as rows. Actions, demonstration targets, and residuals are expressed in the refiner's fixed normalized action coordinates. Auxiliary poses and object trajectories retain physical units. Aegis-VLA applies gated arm residuals while retaining the base gripper actions. Object-state and future robot-state annotations are used only for training.

\subsection{Trajectory World Model}

\begin{figure*}
    \vspace{+0.15cm}
    \centering
    \includegraphics[width=0.99\linewidth]{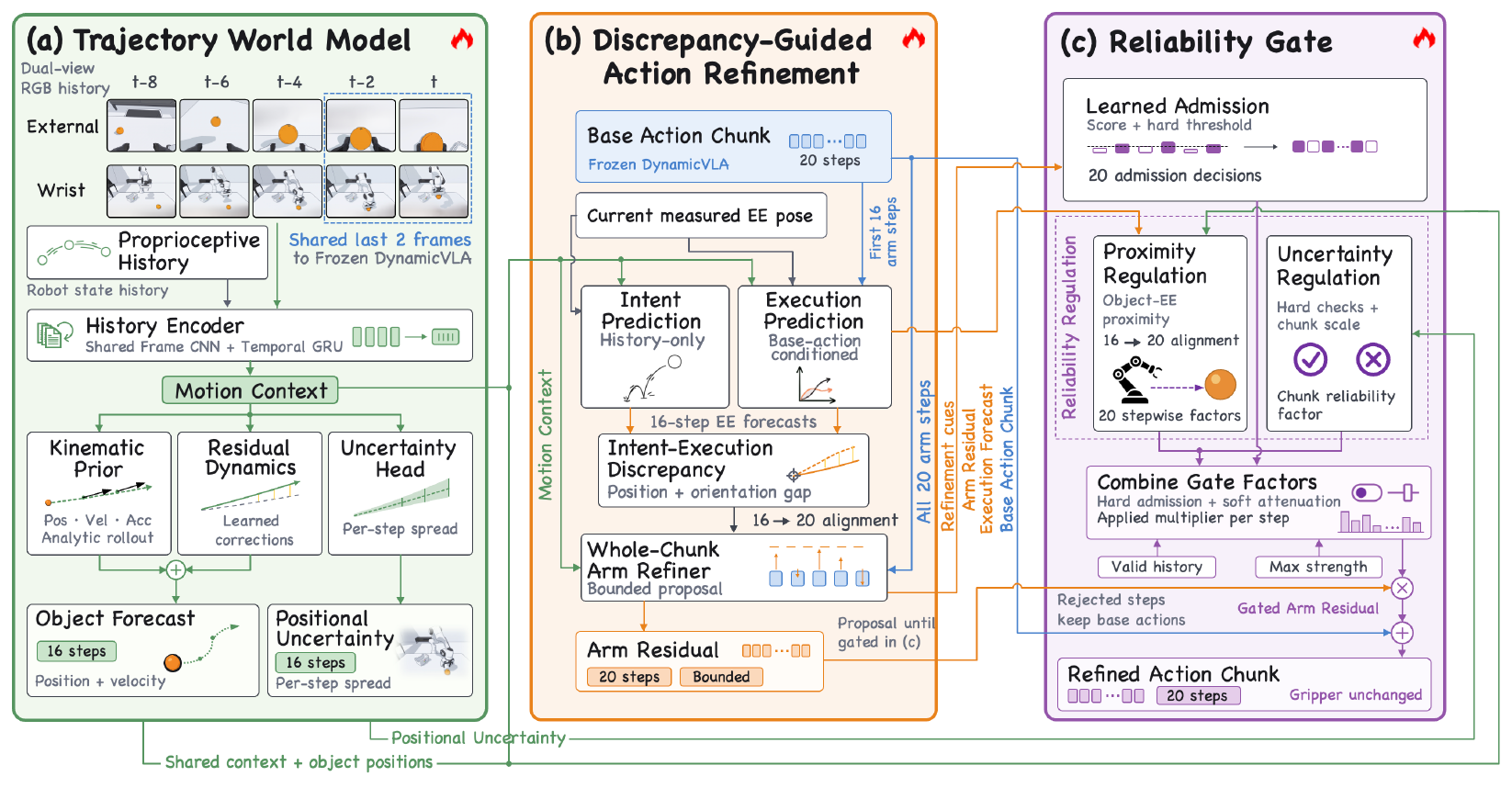}
    \vspace{-0.10cm}
    \caption{Internal architecture of Aegis-VLA. (a) The Trajectory World Model combines kinematic priors and learned residual dynamics to forecast object trajectories and positional uncertainty from dual-view RGB and proprioceptive history. (b) Discrepancy-Guided Action Refinement combines shared motion context, base actions, and discrepancies between history-only and base-action-conditioned end-effector forecasts to propose bounded arm residuals for the full action chunk. (c) The Reliability Gate selects and scales each arm correction using learned admission and uncertainty/proximity regulation.}
    \label{modules}
    \vspace{-0.45cm}
\end{figure*}

As illustrated in Fig.~\ref{modules}(a), a convolutional encoder shared across the two views and history entries, followed by a temporal gated recurrent unit (GRU), maps $\mathcal H_t$ to $\mathbf z_t\in\mathbb R^{d_{\mathrm{ctx}}}$, where $\mathbf z_t$ is the motion context and $d_{\mathrm{ctx}}$ is its feature dimension. The forecasting, refinement, and gating branches share the resulting motion context.
From this context, a state head estimates current object kinematics for constant-acceleration extrapolation, while a residual dynamics head learns corrections for departures from this motion prior:
\begin{equation}
\begin{aligned}
\hat{\mathbf p}_{t,h}
&=\hat{\mathbf p}_{t}+\hat{\mathbf v}_{t}\delta_h
  +\tfrac12\hat{\mathbf a}_{t}\delta_h^2
  +\boldsymbol\rho_{t,h}^{\mathrm{pos}}
  ,\\
\hat{\mathbf v}_{t,h}
&=\hat{\mathbf v}_{t}+\hat{\mathbf a}_{t}\delta_h
  +\boldsymbol\rho_{t,h}^{\mathrm{vel}},
\end{aligned}
\label{eq:aegis_kinematics}
\end{equation}
where $\hat{\mathbf p}_{t}$, $\hat{\mathbf v}_{t}$, and $\hat{\mathbf a}_{t}\in\mathbb R^3$ are the predicted current object position, velocity, and acceleration, 
measured in $\mathrm{m}$, $\mathrm{m}\,\mathrm{s}^{-1}$, and $\mathrm{m}\,\mathrm{s}^{-2}$, respectively.
The index $h\in\{1,\ldots,M\}$ enumerates $M$ future steps, $\delta_h=h/f_{\mathrm{s}}$ is the future offset in seconds, and $\hat{\mathbf p}_{t,h},\hat{\mathbf v}_{t,h}\in\mathbb R^3$ are the corresponding object forecasts. The learned vectors $\boldsymbol\rho_{t,h}^{\mathrm{pos}},\boldsymbol\rho_{t,h}^{\mathrm{vel}}\in\mathbb R^3$ 
correct position and velocity in their respective units. In our experiments, forecasts span $0.64\,\mathrm{s}$ over $M=16$ steps at $f_{\mathrm{s}}=25\,\mathrm{Hz}$, with distinct learned position and velocity corrections.

A separate head maps $\mathbf z_t$ to bounded log-standard deviations, yielding $\sigma_{t,h}\in\mathbb R_{>0}$, where $\sigma_{t,h}$ is a predicted positional standard deviation in meters. At each forecast step, a single scalar is shared across the three spatial coordinates. We learn this input-dependent error scale with a heteroscedastic Gaussian negative log-likelihood:
\begin{equation}
\mathcal L_{t,h}^{\mathrm{unc}}
=\frac{\|\hat{\mathbf p}_{t,h}-\mathbf p_{t,h}^{\star}\|_2^2}
        {6\sigma_{t,h}^{2}}+\log\sigma_{t,h},
\label{eq:aegis_uncertainty}
\end{equation}
where $\mathcal L_{t,h}^{\mathrm{unc}}$ is the per-step uncertainty loss, and $\mathbf p_{t,h}^{\star}\in\mathbb R^3$ is the annotated object position at observation index $t+h$. The factor $6$ follows from averaging the Gaussian quadratic term over three coordinates, with an additive constant omitted. 

Current-state and future position/velocity supervision, including losses for errors along and across the target's motion, shapes the motion context used for refinement. Explicit object positions enable proximity prediction in the Reliability Gate, while positional uncertainty regulates the application of proposed corrections.

\subsection{Discrepancy-Guided Action Refinement}
\label{sec:aegis_refinement}

Fig.~\ref{modules}(b) shows how Aegis-VLA relates its predictive context to the base action. A history-only multilayer perceptron (MLP) predicts future end-effector poses $\hat{\mathbf e}_{t,h}^{\mathrm{hist}}\in\mathbb R^6$, while a second GRU additionally conditioned on the first $M$ base arm actions predicts future end-effector poses $\hat{\mathbf e}_{t,h}^{\mathrm{cond}}\in\mathbb R^6$. Both add predicted offsets to the measured current pose $\mathbf s_t$ and are supervised by the same demonstrated future end-effector trajectory.
Their discrepancy provides an action-dependent cue for proposing corrections before execution.

We compute their component-wise discrepancy and map it to the action-chunk indices:
\begin{equation}
\begin{aligned}
\mathbf d_{t,h}
&=\operatorname{wrap}\!\left(
\hat{\mathbf e}_{t,h}^{\mathrm{hist}}
-\hat{\mathbf e}_{t,h}^{\mathrm{cond}}\right),\\
\widetilde{\mathbf D}_{t}
&=
\mathcal T \!\left([\mathbf d_{t,h}]_{h=1}^{M}\right),
\end{aligned}
\label{eq:aegis_discrepancy}
\end{equation}
where $\operatorname{wrap}(\cdot)$ applies the identity to the first three components, which correspond to position, and wraps each of the last three Euler-angle components to $[-\pi,\pi]$, with $\pi$ denoting the circle constant. The operator $\mathcal T(\cdot)$ performs endpoint-aligned linear interpolation along sequence indices, and $\widetilde{\mathbf D}_{t}\in\mathbb R^{K\times6}$ contains the resulting action-indexed features. This operation resamples the discrepancy sequence while preserving the physical forecast horizon.

A shared MLP generates all arm residuals in one batch:
\begin{equation}
\Delta\mathbf u_{t,i}
=\zeta_{\mathrm{max}}\tanh\!\left(
\mathcal F_{\mathrm{ref}}\!\left(
[\mathbf z_t;\mathbf u_{t,i}^{\mathrm{base}};
\tilde{\mathbf d}_{t,i}]\right)\right),
\label{eq:aegis_residual}
\end{equation}
where $\Delta\mathbf u_{t,i}\in\mathbb R^6$ is the proposed arm residual, $\zeta_{\mathrm{max}}>0$ bounds each component in normalized action units, and $\tanh(\cdot)$ acts component-wise. 
$\mathcal F_{\mathrm{ref}}(\cdot):\mathbb R^{d_{\mathrm{ctx}}+12}\rightarrow\mathbb R^6$ is the trainable refiner, taking the concatenated motion context, base arm action, and aligned discrepancy as input. $\tilde{\mathbf d}_{t,i}\in\mathbb R^6$ is the $i$-th row of $\widetilde{\mathbf D}_{t}$. The ungated candidate $\mathbf u_{t,i}^{\mathrm{cand}}=\mathbf u_{t,i}^{\mathrm{base}}+\Delta\mathbf u_{t,i}$ is supervised against the demonstrated arm action. First- and second-difference penalties regularize temporal variation in the residual sequence.


\begin{table*}[t]
\centering
\vspace{+0.15cm}
\caption{Performance on the Dynamic Object Manipulation (DOM) simulation benchmark. Average success rates (Succ Rate, \%) are reported across
nine evaluation sub-dimensions, organized under three categories: Interaction, Perception, and Generalization. In addition,
overall Succ Rate (\%), path length (Path Len, meters), and mean nominal episode duration (Time, seconds) are reported. 
Each method is evaluated over 90 configurations (10 scenes × 9 dimensions) with 20 trials each, totaling 1,800 evaluation trials.
All baseline models are fine-tuned on the DOM dataset
using their official implementations and released pretrained weights. 
\textbf{Bold} and \underline{underlined} values indicate the best and second-best reported results in each column, respectively. Ties share the same formatting.}
\label{tab:dom_main_results}
\begingroup

\providecolor{mypurple}{RGB}{232,226,244}

\newcolumntype{Z}{c}
\newcolumntype{M}{c}
\newcolumntype{V}{c}
\newcolumntype{S}{c}
\newcolumntype{P}{c}
\newcolumntype{T}{c}
\setlength{\tabcolsep}{1.2pt}
\renewcommand{\arraystretch}{1.18}

\resizebox{0.99\textwidth}{!}{%
\fontsize{7.5}{7.5}\selectfont
\begin{tabular}{l !{\vrule width 0.8pt}
Z|Z|Z !{\vrule width 0.8pt}
Z|Z|Z !{\vrule width 0.8pt}
Z|Z|Z !{\vrule width 0.8pt}
S|T|P}
\toprule[1.5pt]
 & \multicolumn{3}{c !{\vrule width 0.8pt}}{\textbf{Interaction}}
 & \multicolumn{3}{c !{\vrule width 0.8pt}}{\textbf{Perception}}
 & \multicolumn{3}{c !{\vrule width 0.8pt}}{\textbf{Generalization}}
 & \multicolumn{3}{c}{\textbf{Average}} \\
\cmidrule(l{2pt}r{2pt}){2-4}
\cmidrule(l{2pt}r{2pt}){5-7}
\cmidrule(l{2pt}r{2pt}){8-10}
\cmidrule(l{2pt}r{2pt}){11-13}
\textbf{Methods}
 & \textbf{CR} & \textbf{DA} & \textbf{LS}
 & \textbf{VU} & \textbf{SR} & \textbf{MP}
 & \textbf{VG} & \textbf{MG} & \textbf{DR}
 & \textbf{Succ Rate} $\uparrow$ & \textbf{Path Len} $\downarrow$ & \textbf{Time} $\downarrow$ \\
\midrule[1pt]
Diffusion Policy~\cite{diffusionpolicy} & 0.50 & 0.50 & 0.00 & 1.00 & 0.00 & 0.00 & 1.00 & 0.50 & 0.00 & 0.38 & 1.34 & 10.89 \\
OpenVLA-OFT~\cite{openvlaoft} & 3.50 & 0.50 & 0.50 & 0.00 & 1.50 & 0.50 & 3.50 & 2.00 & 0.00 & 1.33 & \textbf{1.08} & 10.83 \\
$\pi_{0}$~\cite{pizero} & 7.50 & 12.00 & 3.00 & 5.50 & 10.50 & 7.50 & 5.50 & 12.50 & 9.00 & 8.11 & \underline{1.19} & 10.55 \\
$\pi_{0.5}$~\cite{pizerofive} & 9.50 & 17.50 & 3.50 & 5.00 & 12.50 & 9.00 & 5.00 & 19.50 & 18.00 & 11.06 & 1.28 & 10.62 \\
SmolVLA~\cite{smolvla} & 18.50 & 17.50 & 5.50 & 1.50 & 14.50 & 11.50 & 14.50 & 13.50 & 17.00 & 12.67 & 1.30 & 10.65 \\
GR00T-N1.5~\cite{groot} & 10.50 & 12.00 & 4.00 & 9.50 & 13.50 & 14.00 & 14.50 & 19.50 & 20.00 & 13.05 & 1.29 & 10.56 \\
VLA-Adapter-Pro~\cite{vlaadapter} & 21.00 & 15.50 & 6.00 & 6.50 & 16.50 & 10.50 & 15.00 & 18.50 & 13.00 & 13.61 & 1.51 & \underline{9.98} \\
VLASH~\cite{vlash} & 9.00 & 20.50 & \underline{7.50} & 6.50 & 7.50 & 12.00 & 7.00 & 21.00 & 20.00 & 12.33 & 1.27 & 10.60 \\
DynamicVLA~\cite{dynamicvla} & \underline{62.00} & \underline{32.50} & \textbf{34.50} & \underline{41.50} & \underline{40.50} & \underline{38.00} & \underline{62.50} & \underline{73.50} & \underline{24.00} & \underline{45.44} & 2.41 & \textbf{8.43} \\
\midrule[1pt]
\rowcolor{mypurple}
\textbf{Aegis-VLA (Ours)} & \textbf{72.00} & \textbf{59.50} & \textbf{34.50} & \textbf{44.00} & \textbf{53.50} & \textbf{63.50} & \textbf{69.50} & \textbf{80.00} & \textbf{34.00} & \textbf{56.72} & 1.87 & \textbf{8.43} \\
\bottomrule[1.5pt]
\end{tabular}%
}
\par\vspace{6pt}
\parbox{\textwidth}{\footnotesize
\textit{Notes:}
CR: Closed-loop Reactivity; DA: Dynamic Adaptation;
LS: Long-horizon Sequencing; VU: Visual Understanding;
SR under Perception: Spatial Reasoning; MP: Motion Perception;
VG: Visual Generalization; MG: Motion Generalization;
DR: Disturbance Robustness.
}
\endgroup
\vspace{-0.35cm}
\end{table*}

\subsection{Reliability Gate}
\label{sec:aegis_gate}

The Reliability Gate in Fig.~\ref{modules}(c) combines action-specific admission with uncertainty, proximity, and history-validity constraints. A sigmoid MLP produces $q_{t,i}\in(0,1)$, where $q_{t,i}$ is the selector score for action step $i$. Its inputs comprise the motion context, base arm action, absolute residual components, residual norm, positional discrepancy norm, and relative position within the chunk. 

\paragraph{Learning admission from relative improvement}
For each valid action step, we assess the base and candidate using $E_{t,i}^{\mathrm{base}}$ and $E_{t,i}^{\mathrm{cand}}$, where these quantities denote their respective mean squared errors against the demonstrated arm action, averaged over the six normalized components. 
Their difference $\Delta E_{t,i}=E_{t,i}^{\mathrm{base}}-E_{t,i}^{\mathrm{cand}}$ denotes the candidate's error reduction. The selector target is
\begin{equation}
\begin{aligned}
y_{t,i}
={}&\boldsymbol{1} [\Delta E_{t,i}>\epsilon_{\mathrm{abs}}]\\
\cdot&\boldsymbol{1}\!\left[
\frac{\Delta E_{t,i}}
{\max(E_{t,i}^{\mathrm{base}},\epsilon_{\mathrm{den}})}
\ge\lambda_{\mathrm{rel}}\right]\\
\cdot&\boldsymbol{1}[\operatorname{ADE}_{t}\le\epsilon_{\mathrm{ADE}}],
\end{aligned}
\label{eq:aegis_precision_target}
\end{equation}
where $y_{t,i}\in\{0,1\}$ is the binary admission target and $\boldsymbol{1}[\cdot]$ equals one when its condition holds and zero otherwise. The positive constants $\epsilon_{\mathrm{abs}}$ and $\epsilon_{\mathrm{den}}$ are the absolute-improvement margin and denominator floor in squared normalized action units, respectively. $\lambda_{\mathrm{rel}}\in[0,1)$ is the minimum relative improvement. The quantity $\operatorname{ADE}_{t}$ is the mean Euclidean object-position error over valid forecast steps, and $\epsilon_{\mathrm{ADE}}>0$ is its tolerance in meters. 
Every other valid action step receives a negative label, with padded steps excluded. This target evaluates the full ungated proposal using demonstration error and forecast accuracy, rather than task-success feedback.

The selector is trained with binary cross-entropy using increased weight on negative labels, and a preservation penalty discourages residual energy on negative steps. Labels are detached from the computation graph. The admission score consequently estimates compliance with these supervised criteria for each proposed correction rather than serving as a calibrated probability of task success.


\paragraph{Reliability-constrained application}
We summarize predictive uncertainty by $\bar\sigma_t=M^{-1}\sum_{h=1}^{M}\sigma_{t,h}$, where $\bar\sigma_t$ is the arithmetic mean of the predicted positional standard deviations, in meters. A proximity head uses the motion context, predicted object position expressed relative to the base-action-conditioned end-effector forecast, and predicted end-effector orientation. 
It is trained with binary object--end-effector proximity labels indicating whether the Euclidean distance between annotated object and demonstrated end-effector positions at the same future step is below a fixed threshold. Its sigmoid probabilities are interpolated from $M$ to $K$ steps using 
$\mathcal T(\cdot)$, yielding $\gamma_{t,i}\in[0,1]$
as the resulting action-indexed proximity score. The applied scaling coefficient is
\begin{equation}
\begin{aligned}
r_t&=\exp(-\bar\sigma_t/\sigma_0),\\
\beta_{t,i}
&=\chi_t\cdot\boldsymbol{1}[q_{t,i}\ge\tau]
\cdot \boldsymbol{1}[\bar\sigma_t\le\sigma_{\mathrm{veto}},\ r_t\ge r_{\mathrm{min}}],\\
\alpha_{t,i}
&=\alpha_{\mathrm{max}}\,\beta_{t,i}\,r_t(1-\kappa\gamma_{t,i}),
\end{aligned}
\label{eq:aegis_gate_coefficient}
\end{equation}
where $r_t\in(0,1]$ is the chunk-level uncertainty attenuation factor, $\sigma_0>0$ is its scale in meters, and $\exp(\cdot)$ denotes the exponential function. The binary variable $\beta_{t,i}$ denotes admission, $\chi_t\in\{0,1\}$ indicates whether all five required history entries are available, and $\tau\in(0,1)$ is the fixed selector threshold. The constants $\sigma_{\mathrm{veto}}>0$ and $r_{\mathrm{min}}\in[0,1]$ are the maximum permitted mean uncertainty in meters and minimum permitted attenuation factor, respectively. 
The coefficient $\alpha_{t,i}\in[0,\alpha_{\mathrm{max}}]$ scales the residual, $\alpha_{\mathrm{max}}\in(0,1]$ is its maximum strength, and $\kappa\in[0,1]$ controls proximity-dependent suppression.
The proximity factor lies in $[0,1]$ and attenuates admitted corrections to limit additional intervention during close approach.

The final actions are
\begin{equation}
\begin{aligned}
\mathbf u_{t,i}^{\mathrm{out}}
&=\mathbf u_{t,i}^{\mathrm{base}}+\alpha_{t,i}\Delta\mathbf u_{t,i},\\
g_{t,i}^{\mathrm{out}}&=g_{t,i}^{\mathrm{base}},
\end{aligned}
\label{eq:aegis_final_action}
\end{equation}
where $\mathbf u_{t,i}^{\mathrm{out}}\in\mathbb R^6$ and $g_{t,i}^{\mathrm{out}}\in\mathbb R$ are the output arm and gripper actions. A rejected correction corresponds to $\alpha_{t,i}=0$ and retains its corresponding base arm action. Thus, admission jointly depends on the selector decision, history validity, and uncertainty checks, while continuous regulation determines the applied magnitude.

\begin{figure*}[t]
    \vspace{+0.15cm}
    \centering
    \includegraphics[width=\textwidth]{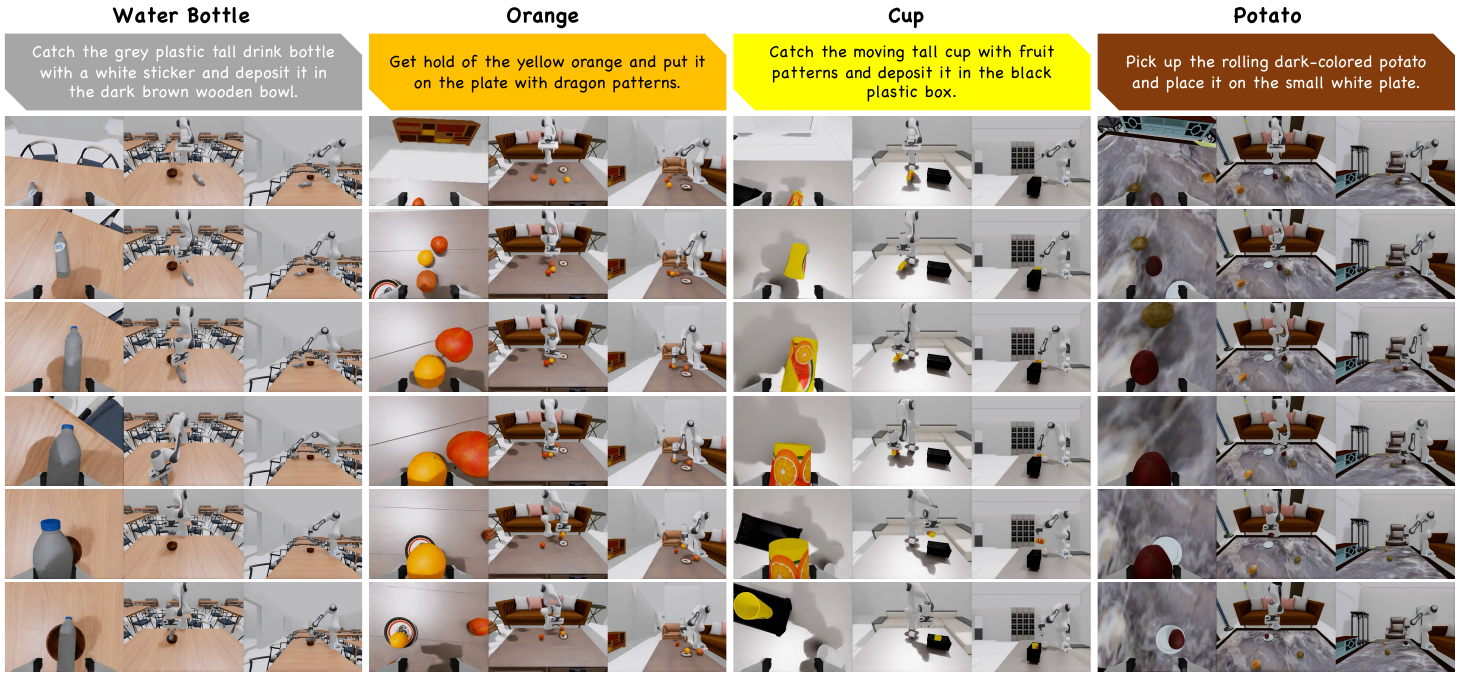}
    \caption{Qualitative Aegis-VLA rollouts on DOM in Isaac Sim, including scenes with multiple distractors. A Franka Emika Panda arm manipulates language-specified moving targets such as a water bottle, an orange, a cup, and a potato. Each column shows one task from multiple camera views, with time progressing downward through approach, grasping, and placement.}
    \label{simulation}
    \vspace{-0.55cm}
\end{figure*}

\section{Experiments}
\label{sec:aegis_experiments}

\subsection{Experimental Setup}
\label{sec:aegis_protocol}

\paragraph{Benchmark and evaluation}
We use the released simulation training data of the Dynamic Object Manipulation (DOM) benchmark~\cite{dynamicvla}, comprising approximately $200{,}000$ demonstrations across $2{,}800$ scenes and $206$ objects. We evaluate Aegis-VLA on DOM using a Franka Emika Panda arm in Isaac Sim. The benchmark covers nine evaluation dimensions with $10$ configurations per dimension and $20$ trials per configuration, totaling $1{,}800$ trials. Evaluation uses the benchmark's task conditions and episode limits, streaming execution, and an evaluator tolerance of $0.07\,\mathrm{m}$. Simulation and policy inference run on two separate RTX 4090 GPUs. We reproduce the DynamicVLA baseline from the released DOM-trained checkpoint.

\paragraph{Model and training}
Aegis-VLA is trained offline using demonstration and auxiliary trajectory supervision. Optimization updates only its forecasting, refinement, and gating modules, comprising approximately $2.45$M parameters, while the action-generation backbone remains frozen.
The backbone generates action chunks of $K=20$ steps.

Training uses AdamW for $80{,}000$ updates on eight A100 GPUs, with a global batch size of $352$ and bfloat16 precision. Learning rates are $5\times10^{-5}$, $10^{-4}$, and $3\times10^{-5}$ for the world/trajectory, refiner, and selector/proximity groups, respectively. At inference, the residual bound is $\zeta_{\mathrm{max}}=0.60$, the maximum correction multiplier is $\alpha_{\mathrm{max}}=0.70$, and the admission threshold is $\tau=0.985$. Gate parameters remain fixed across test configurations.

\paragraph{Evaluation metrics}
Success rate measures completion of the instructed task under the official success and termination conditions. Mean end-effector path length and nominal episode duration include all trials, including failures. Duration is the simulator-step count multiplied by $0.04\,\mathrm{s}$ and does not measure inference latency. Each reported local setting is evaluated in a single complete run. 
Percentage-point differences are based on the displayed success rates.

\subsection{Comparative Results}

\paragraph{Task performance}
Table~\ref{tab:dom_main_results} compares Aegis-VLA with representative DOM baselines. Aegis-VLA achieves $56.72\%$ overall success, improving on the reproduced baseline of $45.44\%$ by $11.28$ percentage points. Success increases in eight of the nine dimensions and matches the baseline on Long-horizon Sequencing ($34.50\%$). The largest gains occur in Dynamic Adaptation, from $32.50\%$ to $59.50\%$, and Motion Perception, from $38.00\%$ to $63.50\%$. Beyond these motion-related dimensions, success also increases in Visual Generalization, Motion Generalization, and Disturbance Robustness by $7.00$, $6.50$, and $10.00$ percentage points, respectively. 

Mean end-effector path length decreases from $2.41$ to $1.87\,\mathrm{m}$, while mean nominal episode duration remains $8.43\,\mathrm{s}$ at the reported precision. Higher success is therefore accompanied by a shorter average path, although these all-trial averages do not isolate efficiency on successful episodes.

\paragraph{Computational efficiency}
We benchmark offline inference on a single RTX~4090 at batch size one with fixed-shape synthetic inputs, native runtime precision, $30$ warm-up queries, and $300$ timed queries per model. Aegis-VLA introduces $2.45$M trainable parameters, equivalent to $0.57\%$ of the backbone's parameter count. Mean latency per $20$-action chunk increases from $227.06$ to $239.31\,\mathrm{ms}$, an overhead of $5.40\%$. Timing covers full policy inference from prepared GPU-resident inputs, including backbone and five-frame world-model encoding, but excludes input preparation, data transfers, communication, and execution.

\paragraph{Qualitative results}
Fig.~\ref{simulation} shows Aegis-VLA rollouts in Isaac Sim across varied objects and scenes, illustrating the robot approaching and grasping moving targets before placing them at the instructed destinations.

\subsection{Ablation Studies}
\label{sec:aegis_ablation}

\paragraph{Refinement without gating}
All ungated variants in Table~\ref{tab:training_module_combinations} are trained independently and apply their own residuals at scale $0.70$ when history is valid. Discrepancy-guided variants without A retain the shared history encoder but remove object-prediction branches and losses. The encoder learns through end-effector and candidate-action objectives, with gate-dependent losses omitted for the refiner-only model. The plain adapter uses the same history-encoder architecture, training budget, optimizer and learning-rate schedule, residual bound, and candidate-action and temporal-difference loss weights as the full model. It predicts residuals from history features and base actions without trajectory predictors, discrepancy inputs, or gating. Its success falls from the baseline's $45.44\%$ to $22.00\%$, showing that its learned corrections harm base-policy performance. The discrepancy-guided refiner instead exceeds the baseline by $2.67$ points under the same ungated execution rule used for the adapter. These contrasting outcomes show that ungated refinement can be effective, but directly applying learned residuals can also damage performance. Adding object forecasting provides a further $2.45$-point gain through supervision of the shared motion context, while execution remains ungated.

\paragraph{Selective refinement}
Full Aegis-VLA exceeds the independently trained ungated predictive refiner by $6.16$ points, showing that learned admission, reliability regulation, and their associated training objectives together improve an already effective refiner. Without object forecasting, the adapted gate $C^{\dagger}$ learns admission from demonstration-relative candidate improvement over the base action. Forecast-accuracy filtering is removed from admission labels and preservation masks. It retains history checks but removes proximity prediction and uncertainty/proximity regulation, applying the fixed scale only to admitted corrections. Adding this adapted gate improves on the refiner alone by $3.39$ points, demonstrating a benefit from learned selection even without object forecasts. Full Aegis-VLA gains another $5.22$ points when object forecasting, forecast-filtered supervision, and uncertainty/proximity regulation are included together.

\paragraph{Inference-time gate ablations}
Table~\ref{tab:inference_gate_ablation} evaluates the same full Aegis-VLA checkpoint without retraining. Replacing its gate with $\alpha_{t,i}=0.70\chi_t$ reduces success to $17.78\%$, below the backbone baseline. With valid history, previously rejected nonzero residuals enter execution and attenuated corrections become stronger. The independently trained ungated predictive refiner reaches $50.56\%$ at the same scale using different learned residuals. Thus, fixed-scale refinement can be effective when learned separately, but imposing that execution rule on full-model residuals severely degrades task performance.

\begin{table}[t]
\vspace{+0.15cm}
\centering
\caption{Independently trained refinement configurations with a frozen backbone. A denotes object forecasting within the Trajectory World Model, B denotes Discrepancy-Guided Action Refinement, and C denotes the Reliability Gate. $C^{\dagger}$ retains learned admission without forecast-accuracy filtering or uncertainty/proximity regulation.}
\label{tab:training_module_combinations}
\normalsize          
\setlength{\tabcolsep}{2.5pt}
\renewcommand{\arraystretch}{1.15}
\begin{tabularx}{\columnwidth}{@{}Xcccr@{}}
\toprule
Training configuration & A & B & C & Success (\%) \\
\midrule
DynamicVLA (reproduced)
& -- & -- & -- & 45.44 \\
Plain residual adapter
& -- & \textit{Plain} & -- & 22.00 \\
\midrule
Discrepancy-guided refiner
& -- & $\checkmark$ & -- & 48.11 \\
Refiner + object forecasting
& $\checkmark$ & $\checkmark$ & -- & 50.56 \\
Refiner + adapted gate
& -- & $\checkmark$ & $C^{\dagger}$ & 51.50 \\
\midrule
Full Aegis-VLA
& $\checkmark$ & $\checkmark$ & $\checkmark$ & \textbf{56.72} \\
\bottomrule
\end{tabularx}
\vspace{-0.55cm}
\end{table}

Bypassing only $\boldsymbol{1}[q_{t,i}\ge\tau]$ in Eq.~\eqref{eq:aegis_gate_coefficient} retains uncertainty attenuation, the veto, the reliability floor, and proximity suppression, but reduces success by $5.39$ points. Bypassing learned admission therefore hurts performance even when uncertainty and proximity constraints remain active.
Removing uncertainty attenuation, the veto, and the reliability floor instead reduces success by $1.50$ points, while admission, proximity suppression, and uncertainty prediction remain active. This smaller observed decrease indicates that uncertainty-based suppression further improves task performance even when learned admission already filters the proposed corrections.

\begin{table}[t]
\vspace{+0.15cm}
\centering
\caption{Inference ablations of the same full Aegis-VLA checkpoint without retraining. All settings retain history-validity checks. $\Delta$ denotes the percentage-point change from the full model, whose result reuses the main evaluation run.}
\label{tab:inference_gate_ablation}
\normalsize          
\setlength{\tabcolsep}{3pt}
\renewcommand{\arraystretch}{1.15}
\begin{tabularx}{\columnwidth}{@{}Xrr@{}}
\toprule
Inference configuration & Success (\%) & $\Delta$ (pp) \\
\midrule
Full-model residuals at fixed scale & 17.78 & $-38.94$ \\
Learned admission bypassed & 51.33 & $-5.39$ \\
Uncertainty modulation removed & 55.22 & $-1.50$ \\
\midrule
Full Aegis-VLA & \textbf{56.72} & $0.00$ \\
\bottomrule
\end{tabularx}
\vspace{-0.45cm}
\end{table}

\section{Discussion}


Freezing parameters does not freeze behavior. The plain adapter and gate-replacement results show that residuals can degrade task performance even when the backbone remains frozen, because they alter the actions actually executed. Independently trained ungated refiners nevertheless outperform the base policy, and selective refinement improves them further. Aegis-VLA therefore separates correction generation from admission, so a proposed correction is not automatically applied. Admission is learned from each candidate's demonstration-relative improvement over its base action, while uncertainty and proximity constrain whether and how strongly the proposed correction is applied.

The unchanged Long-horizon Sequencing score may reflect the temporal scope of refinement. Corrections are generated from short-horizon end-effector forecasts without explicitly modeling task progress or subtask transitions. If later stages remain unsuccessful, better interception alone does not increase full-sequence success.

Demonstration-relative improvement compares candidates with base actions against recorded demonstrations. Actions that differ from the demonstration may still lead to task success, and lower local action error need not improve contact timing or later task outcomes. The success drop when admission is bypassed supports the learned selector's practical benefit, although the task-level results do not establish whether every admitted correction is beneficial.

The current evaluation is simulation-based, and smaller ablation differences warrant repeated runs. Future work will examine selective refinement on physical robots and under shifts in target motion, visibility, and execution dynamics, particularly when predictive evidence becomes less reliable.

\section{Conclusion}
\label{sec:aegis_conclusion}


Aegis-VLA integrates trajectory forecasting, discrepancy-guided arm refinement, and reliability gating for dynamic object manipulation. With approximately $2.45$M trainable parameters, it achieves $56.72\%$ DOM success, improving on the reproduced baseline by $11.28$ percentage points. 
Ablations show that the full framework further improves on an independently trained ungated refiner that already outperforms the base policy. Aegis-VLA thus improves adaptation to moving targets with modest additional parameter and measured inference costs.

\bibliographystyle{ieeetr}
\bibliography{method}

@article{chunk,
  title={Real-time execution of action chunking flow policies},
  author={Black, Kevin and Galliker, Manuel and Levine, Sergey},
  journal={Advances in Neural Information Processing Systems},
  year={2026}
}

@article{a2c2,
  title={Leave no observation behind: Real-time correction for vla action chunks},
  author={Sendai, Kohei and Alvarez, Maxime and Matsushima, Tatsuya and Matsuo, Yutaka and Iwasawa, Yusuke},
  journal={arXiv preprint arXiv:2509.23224},
  year={2025}
}

@article{dynamicvla,
  title={Dynamicvla: A vision-language-action model for dynamic object manipulation},
  author={Xie, Haozhe and Wen, Beichen and Zheng, Jiarui and Chen, Zhaoxi and Hong, Fangzhou and Diao, Haiwen and Liu, Ziwei},
  journal={arXiv preprint arXiv:2601.22153},
  year={2026}
}

@article{puma,
  title={Towards generalizable robotic manipulation in dynamic environments},
  author={Fang, Heng and Li, Shangru and Wang, Shuhan and Xi, Xuanyang and Liang, Dingkang and Bai, Xiang},
  journal={arXiv preprint arXiv:2603.15620},
  year={2026}
}

@article{ahead,
  title={Intercepting the future: Latent-space predictive world model for dynamic VLA manipulation},
  author={Syed, Shahram Najam and Jakobsson, Arthur and Hao, Haoran and Ichnowski, Jeffrey},
  journal={arXiv preprint arXiv:2606.02486},
  year={2026}
}

@article{dynawm,
  title={DynaWM: A Base-VLA-Guided World Foundation Model for Moving-Object Manipulation},
  author={Chang, Chongkei and Deng, Zhidong},
  journal={arXiv preprint arXiv:2607.02604},
  year={2026}
}

@misc{dreamtrajectory,
  author = {Yang, Zheng and Zhang, Wenjie and Chen, Xiangyu
            and Song, Wenxuan and Wang, Xianpeng and Kang, Yihang
            and Wen, Jiawen and Chen, Wen and Wang, Lujia
            and Xu, Renjing and Li, Haoang and Chu, Xiaowen},
  title = {{DreamTrajectory}: Trajectory-Guided Action Generation
           with World Model Alignment for Mobile Manipulation},
  year = {2026},
  howpublished = {arXiv:2608.01381},
  url = {https://arxiv.org/abs/2608.01381v2}
}

@misc{pizero,
      title={$\pi_0$: A Vision-Language-Action Flow Model for General Robot Control}, 
      author={Kevin Black and Noah Brown and Danny Driess and Adnan Esmail and Michael Equi and Chelsea Finn and Niccolo Fusai and Lachy Groom and Karol Hausman and Brian Ichter and Szymon Jakubczak and Tim Jones and Liyiming Ke and Sergey Levine and Adrian Li-Bell and Mohith Mothukuri and Suraj Nair and Karl Pertsch and Lucy Xiaoyang Shi and James Tanner and Quan Vuong and Anna Walling and Haohuan Wang and Ury Zhilinsky},
      year={2026},
      archivePrefix={arXiv},
      url={https://arxiv.org/abs/2410.24164}, 
}

@misc{pizerofive,
      title={$\pi_{0.5}$: a Vision-Language-Action Model with Open-World Generalization}, 
      author={Physical Intelligence and Kevin Black and Noah Brown and James Darpinian and Karan Dhabalia and Danny Driess and Adnan Esmail and Michael Equi and Chelsea Finn and Niccolo Fusai and Manuel Y. Galliker and Dibya Ghosh and Lachy Groom and Karol Hausman and Brian Ichter and Szymon Jakubczak and Tim Jones and Liyiming Ke and Devin LeBlanc and Sergey Levine and Adrian Li-Bell and Mohith Mothukuri and Suraj Nair and Karl Pertsch and Allen Z. Ren and Lucy Xiaoyang Shi and Laura Smith and Jost Tobias Springenberg and Kyle Stachowicz and James Tanner and Quan Vuong and Homer Walke and Anna Walling and Haohuan Wang and Lili Yu and Ury Zhilinsky},
      year={2025},
      eprint={2504.16054},
      archivePrefix={arXiv},
      primaryClass={cs.LG},
      url={https://arxiv.org/abs/2504.16054}, 
}

@article{dynamicwam,
  title={DynamicWAM: Dual-Path Motion Conditioning for World-Action Models in Dynamic Manipulation},
  author={Lou, Yunfan and Gao, Hewen and Zhu, Xiyu and Qiao, Zhuoran and Han, Xuan and Yang, Yifan and Ye, Yifan and Yao, Boxian and Pang, Zhibo},
  journal={arXiv preprint arXiv:2608.00793},
  year={2026}
}

@article{rt2,
  title={Rt-2: Vision-language-action models transfer web knowledge to robotic control},
  author={Brohan, Anthony and Brown, Noah and Carbajal, Justice and Chebotar, Yevgen and Chen, Xi and Choromanski, Krzysztof and Ding, Tianli and Driess, Danny and Dubey, Avinava and Finn, Chelsea and others},
  journal={arXiv preprint arXiv:2307.15818},
  year={2023}
}

@article{openvla,
  title={Openvla: An open-source vision-language-action model},
  author={Kim, Moo Jin and Pertsch, Karl and Karamcheti, Siddharth and Xiao, Ted and Balakrishna, Ashwin and Nair, Suraj and Rafailov, Rafael and Foster, Ethan and Lam, Grace and Sanketi, Pannag and others},
  journal={arXiv preprint arXiv:2406.09246},
  year={2024}
}

@inproceedings{tracevla,
  title={Tracevla: Visual trace prompting enhances spatial-temporal awareness for generalist robotic policies},
  author={Zheng, Ruijie and Liang, Yongyuan and Huang, Shuaiyi and Gao, Jianfeng and Daum{\'e} III, Hal and Kolobov, Andrey and Huang, Furong and Yang, Jianwei},
  booktitle={International Conference on Learning Representations},
  year={2025}
}

@article{smolvla,
  title={Smolvla: A vision-language-action model for affordable and efficient robotics},
  author={Shukor, Mustafa and Aubakirova, Dana and Capuano, Francesco and Kooijmans, Pepijn and Palma, Steven and Zouitine, Adil and Aractingi, Michel and Pascal, Caroline and Russi, Martino and Marafioti, Andres and others},
  journal={arXiv preprint arXiv:2506.01844},
  year={2025}
}

@article{vlash,
  title={Vlash: Real-time vlas via future-state-aware asynchronous inference},
  author={Tang, Jiaming and Sun, Yufei and Zhao, Yilong and Yang, Shang and Lin, Yujun and Zhang, Zhuoyang and Hou, James and Lu, Yao and Liu, Zhijian and Han, Song},
  journal={arXiv preprint arXiv:2512.01031},
  year={2025}
}

@article{dbctfp,
  title={Dynamic behavior cloning with temporal feature prediction: Enhancing robotic arm manipulation in moving object tasks},
  author={Zhang, Yifan and Wang, Ruiping and Chen, Xilin},
  journal={IEEE Robotics and Automation Letters},
  volume={10},
  number={6},
  pages={5209--5216},
  year={2025},
  publisher={IEEE}
}

@article{gem,
  title={Train once, deploy anywhere: Realize data-efficient dynamic object manipulation},
  author={Li, Zhuoling and Wu, Xiaoyang and Xu, Zhenhua and Zhao, Hengshuang},
  journal={arXiv e-prints},
  year={2025}
}

@inproceedings{gr1,
  title={Unleashing large-scale video generative pre-training for visual robot manipulation},
  author={Wu, Hongtao and Jing, Ya and Cheang, Chilam and Chen, Guangzeng and Xu, Jiafeng and Li, Xinghang and Liu, Minghuan and Li, Hang and Kong, Tao},
  booktitle={International Conference on Learning Representations},
  year={2024}
}

@article{vpp,
  title={Video prediction policy: A generalist robot policy with predictive visual representations},
  author={Hu, Yucheng and Guo, Yanjiang and Wang, Pengchao and Chen, Xiaoyu and Wang, Yen-Jen and Zhang, Jianke and Sreenath, Koushil and Lu, Chaochao and Chen, Jianyu},
  journal={arXiv preprint arXiv:2412.14803},
  year={2024}
}

@article{worldvla,
  title={Worldvla: Towards autoregressive action world model},
  author={Cen, Jun and Yu, Chaohui and Yuan, Hangjie and Jiang, Yuming and Huang, Siteng and Guo, Jiayan and Li, Xin and Song, Yibing and Luo, Hao and Wang, Fan and others},
  journal={arXiv preprint arXiv:2506.21539},
  year={2025}
}

@inproceedings{residualrl,
  title={Residual reinforcement learning for robot control},
  author={Johannink, Tobias and Bahl, Shikhar and Nair, Ashvin and Luo, Jianlan and Kumar, Avinash and Loskyll, Matthias and Ojea, Juan Aparicio and Solowjow, Eugen and Levine, Sergey},
  booktitle={2019 international conference on robotics and automation (ICRA)},
  year={2019},
  organization={IEEE}
}

@inproceedings{policydecorator,
  title={Policy decorator: Model-agnostic online refinement for large policy model},
  author={Yuan, Xiu and Mu, Tongzhou and Tao, Stone and Fang, Yunhao and Zhang, Zhang and Su, Hao},
  booktitle={International Conference on Learning Representations},
  year={2025}
}

@article{uncertaintyresidual,
  title={Accelerating residual reinforcement learning with uncertainty estimation},
  author={Dodeja, Lakshita and Schmeckpeper, Karl and Vats, Shivam and Weng, Thomas and Jia, Mingxi and Konidaris, George and Tellex, Stefanie},
  journal={IEEE Robotics and Automation Letters},
  year={2025},
  publisher={IEEE}
}

@inproceedings{genieenvisioner,
  title={Genie envisioner: A unified world foundation platform for robotic manipulation},
  author={Liao, Yue and Zhou, Pengfei and Huang, Siyuan and Yang, Donglin and Chen, Shengcong and Jiang, Yuxin and Hu, Yue and Liu, Si and Luo, Jianlan and Chen, Liliang and others},
  booktitle={International Conference on Learning Representations},
  year={2026}
}

@inproceedings{resip,
  title={From imitation to refinement-residual rl for precise assembly},
  author={Ankile, Lars and Simeonov, Anthony and Shenfeld, Idan and Torne, Marcel and Agrawal, Pulkit},
  booktitle={2025 IEEE International Conference on Robotics and Automation (ICRA)},
  year={2025},
  organization={IEEE}
}

@article{unipi,
  title={Learning universal policies via text-guided video generation},
  author={Du, Yilun and Yang, Sherry and Dai, Bo and Dai, Hanjun and Nachum, Ofir and Tenenbaum, Josh and Schuurmans, Dale and Abbeel, Pieter},
  journal={Advances in neural information processing systems},
  year={2023}
}

@inproceedings{hifvla,
  title={Hif-vla: Hindsight, insight and foresight through motion representation for vision-language-action models},
  author={Lin, Minghui and Ding, Pengxiang and Wang, Shu and Zhuang, Zifeng and Liu, Yang and Tong, Xinyang and Song, Wenxuan and Lyu, Shangke and Huang, Siteng and Wang, Donglin},
  booktitle={Proceedings of the IEEE/CVF Conference on Computer Vision and Pattern Recognition},
  year={2026}
}

@article{rtvla,
  title={Running vlas at real-time speed},
  author={Ma, Yunchao and Zhou, Yizhuang and Yang, Yunhuan and Wang, Tiancai and Fan, Haoqiang},
  journal={arXiv preprint arXiv:2510.26742},
  year={2025}
}

@article{contextvla,
  title={Contextvla: Vision-language-action model with amortized multi-frame context},
  author={Jang, Huiwon and Yu, Sihyun and Kwon, Heeseung and Jeon, Hojin and Seo, Younggyo and Shin, Jinwoo},
  journal={arXiv preprint arXiv:2510.04246},
  year={2025}
}

@article{openvlaoft,
  title={Fine-tuning vision-language-action models: Optimizing speed and success},
  author={Kim, Moo Jin and Finn, Chelsea and Liang, Percy},
  journal={arXiv preprint arXiv:2502.19645},
  year={2025}
}

@article{diffusionpolicy,
  title={Diffusion policy: Visuomotor policy learning via action diffusion},
  author={Chi, Cheng and Xu, Zhenjia and Feng, Siyuan and Cousineau, Eric and Du, Yilun and Burchfiel, Benjamin and Tedrake, Russ and Song, Shuran},
  journal={The International Journal of Robotics Research},
  year={2025},
  publisher={Sage Publications Sage UK: London, England}
}

@article{act,
  title={Learning fine-grained bimanual manipulation with low-cost hardware},
  author={Zhao, Tony Z and Kumar, Vikash and Levine, Sergey and Finn, Chelsea},
  journal={arXiv preprint arXiv:2304.13705},
  year={2023}
}

@article{octo,
  title={Octo: An open-source generalist robot policy},
  author={Team, Octo Model and Ghosh, Dibya and Walke, Homer and Pertsch, Karl and Black, Kevin and Mees, Oier and Dasari, Sudeep and Hejna, Joey and Kreiman, Tobias and Xu, Charles and others},
  journal={arXiv preprint arXiv:2405.12213},
  year={2024}
}

@article{groot,
  title={Gr00t n1: An open foundation model for generalist humanoid robots},
  author={Bjorck, Johan and Casta{\~n}eda, Fernando and Cherniadev, Nikita and Da, Xingye and Ding, Runyu and Fan, Linxi and Fang, Yu and Fox, Dieter and Hu, Fengyuan and Huang, Spencer and others},
  journal={arXiv preprint arXiv:2503.14734},
  year={2025}
}

@inproceedings{vlaadapter,
  title={Vla-adapter: An effective paradigm for tiny-scale vision-language-action model},
  author={Wang, Yihao and Ding, Pengxiang and Li, Lingxiao and Cui, Can and Ge, Zirui and Tong, Xinyang and Song, Wenxuan and Zhao, Han and Zhao, Wei and Hou, Pengxu and others},
  booktitle={Proceedings of the AAAI conference on artificial intelligence},
  year={2026}
}

\end{document}